# Brain-Like Stochastic Search:
# A Research Challenge and Funding Opportunity

**Paul J. Werbos**
Room 675, NSF[*]
Arlington VA 22230
pwerbos@nsf.gov

Abstract – Brain-Like Stochastic Search (BLiSS) refers to this task: given a <u>family</u> of utility functions U(**u**,α), where α is a vector of parameters or task descriptors, maximize or minimize U with respect to **u**, using networks (Option Nets) which <u>input</u> α and <u>learn</u> to generate good options **u** stochastically. This paper discusses why this is crucial to brain-like intelligence (an area funded by NSF) and to many applications, and discusses various possibilities for network design and training.

## 1   Introduction

Evolutionary computing is probably the lead technology today for finding global minima or maxima to a function U(**u**). Of course, there are many forms of evolutionary computing. There are also classical methods, like Gibbs search and the sophisticated trust region approaches recently developed by Barhen et al and used on the Desert Storm tank routing problem. There are a few neural net designs (like Kohonen nets, but not Hopfield nets) which have had competitive performance on some specialized large-scale optimization problems.

On the other had, it is hard to believe that the human brain uses these kinds of algorithms directly in making complex, novel decisions. As a result, many people doing basic research in neural networks have essentially ignored the need for this kind of systematic stochastic search. Some kinds of stochastic exploration methods have been developed (e.g., see Thrun in White and Sofge, 1992), but nothing with the kind of complexity and richness one finds in the evolutionary computation literature. The neural network field, in turn, still pays a lead role in developing the kind of highly functional network models which, in my view, are the only models which have a serious chance of eventually explaining the functional power of biological neural networks. (See Werbos in Pribram 1998.)

More recent work on neural networks suggests that this is a major omission – that greater attention to stochastic search will be a necessary part of replicating or understanding the higher-order intelligence that we find even in the lowest of mammal brains. Because this goal is a major goal for research sponsored at NSF, we are changing our priorities in the Control, Network and Computational Intelligence (CNCI) program to encourage research in this topic, and to encourage several related topics. Section 2 of this paper will discuss those other, related topics, and will discuss my general reasoning here.

From a practical viewpoint, BLiSS offers a number of obvious advantage. In essence, the training of Option Nets over a variety of problems in a given domain allows you (your network) to build up domain-specific knowledge about how to solve optimization problems in a given <u>domain</u>. For example, if you define α to be the set of coordinates of 100 cities in the 100-city Traveling Salesman Problem (TSP), an Option net could learn how to set up good initial guesses for the optimum. Perhaps a recurrent Option Net (taking the previous best guess as an input) could make good guesses about how to <u>improve</u> the initial guesses. It would be interesting to see how training could improve the strength of such a search mechanism, and compare with other less domain-specific search methods. as another example, there would be great value in training a system specifically to solve problems in VLSI design, or to solve problems in aircraft design, and so on.

In a brain-like context, the system somehow needs to search over <u>millions</u> of choices, not just the dozen or two which are most comon in engineering today. In order to handle such very large problems, there is a need for the system to learn <u>which variables</u> to focus on in the search. Generic search techniques, which do not incorporate the learning of where to focus, cannot ultimately handle such large problems.

Of course, there has been a little work in the evolutionary computing area on ways to tweak parameters of an evolutionary search, analogous to tweaking the learning rates in neural networks. But the goal here is to achieve a more all-encompassing sort of learning, which, once again, has some serious hope of explaining how the human brain handles these kinds of problems.

The next section will say more about the concept of brain-like stochastic search, neural networks, and the broader goals of the CNCI program. Those who really do not care about that background may want to jump ahead to sections 3 and 4 which discuss the training and structure of Option Nets and Bliss Systems. **Please note that the suggestion in sections 3 and 4 are very tentative**; they are only my attempt to indicate that something like BLiSS is possible, and to suggest a couple of possible ways to get started in such work. NSF strongly encourages any innovative approach whatsoever that seriously addresses the underlying research challenge here; there is absolutely no

---

intention to give preference to the examples which I happen to be able to think of myself at this moment. It is strongly hoped that your community, because of its unique background and way of thinking, will be able to develop new approaches, to fill in this critical blank spot in our present knowledge, where there is hardly any existing literature to compete with.

## 2   BLiSS and Brain-Like Intelligence

Many years ago, D.O. Hebb – one of the grandfathers of the neural network field – proposed that higher order intelligence in the brain results as a kind of emergent phenomenon. He proposed that we could replicate that kind of intelligence, simply by discovering the right "general neuron model," including the equations to adapt that general neuron. he proposed that we could develop an artificial intelligent system, simply by hooking together trillions of such model neurons, hooking them up to sensor input and motor output, and letting them learn everything from experience.

From an engineering viewpoint, there are reasons to doubt that this could be done, using Hebb's ideas about learning, and one type of neuron only. However, for my Harvard PhD thesis proposal, back in 1972, I suggested that we might achieve Hebb's dream, by using three types of neuron – one to implement a "critic" network, one to learn a dynamic model of the external world, and one to execute actions. (See Werbos in Pribram 1994, and Werbos 1994.) Backpropagation was part of this design – as was the very first reinforcement learning design linked to dynamic programming. (See Werbos 1998a.)

As late as 1992 or 1994, this seemed like a reasonable goal. In any event, there are reasons to believe that neural network designs in this family can outperform traditional methods for intelligent control in a wide variety of applications (Werbos 1999) and in terms of rigorous stability results (Werbos 1998a).

Based on this general vision, and based on the views of control theory, I proposed that we could achieve (mammal) brain-like intelligence by advancing and coupling basic research in three critical topics: (1) improved supervised learning (neural or nonneural), which provides the basic building blocks for more complex systems; (2) learning based system identification, which provides the necessary prediction network; (3) approximate dynamic programming (in relation to other control methods), which could provide the overall architecture of an intelligent system – including the critic and action components.

This kind of design has done better and better in practical engineering applications, but the gap between this approach and the capabilities of the brain has become more and more apparent. In a long paper cited in Werbos (1999), I summarize what the nature of the gap is. First, many artificial intelligence (AI) people have long argued that there is a difference between control and decision. For example, the problem of how to move your muscles at every moment is problem in control. But the problem of deciding where to go to college is a decision problem – it conditions your actions and life for years to come. Originally, we hoped that a good capability at making decisions would emerge from a simpler architecture, without a need to impose any kind of hierarchy or stratified system of controls and decisions. But recent work, both in biology and in analysis of learning methods, suggests that this approach is false. I have developed (and applied for a patent on) a generalized Bellman equation, considerably more general and flexible than some alternative approaches suggested by Sutton, which provides a starting point for the design of such a hierarchical learning-based decision-making system. In general, I would consider the area of "temporal chunking" (decision making as opposed to control in the narrow sense) to be one of the four new topics which requires more research – interdisciplinary research combining the best of what is known, at least, in AI and neural networks. I would also regard the idea of "multiple models" to be one possible testbed or formulation of this research task.

Related to this task is the area of spatial chunking, which I will discuss further below. This essentially involves three sets of similar tasks: (1) the exploitation of symmetry or "object structure" to construct neural networks which do not have fixed numbers of inputs and outputs; (2) true spatial chunking – as in creating condensed roadmaps from complex aerial views; (3) world modeling – the management of decision problems when the world we see (even including some filtering and picturing of nearby objects out of sight) is actually a very small part of the larger world we want to cope with.

Now: where does the third new task, BLiSS, come into it? Very simply – the problem of local minima versus global minima starts to become overwhelming when we start to think in terms of decisions as opposed to control. For example, we can easily use traditional control methods to control our hand, to make it place a "Go" piece onto the "Go" board, to a desired location. But in deciding WHERE to move, we basically face 361 local minima. (each grid point on a 19 by 19 board.) We desperately need options.

Biologically (Werbos in Pribram 1998), the commitment to a decision seems to take place in an area called the basal ganglia, which was once very mysterious but is now becoming better understood. The development of options seems to take place at a layer of the higher cerebral cortex (layer V or VI, I forget which), which is in fact known to have some stochastic behavior. It is also fascinating to consider that our image of reality may also be stochastic, may be constructed as a kind of decision by this layer of the brain. In fact, Bitterman showed long ago that the ability to handle certain stochastic aspects of reality is the hallmark of the mammal level of intelligence, compared to lower classes of vertebrate.

A fourth "new" topic is the training of neural nets to represent probability distributions, or probability distributions conditional on past information. Actually, his was always a theoretical priority; I would view the training of stochastic models of the world as part of the topic of system identification. (e.g. see Chapter 13 of White and Sofge, 192.) However, since this priority did not receive serious attention from the university research community in that formulation,. I would add a new topic description, to try to suggest a unification to the many various strands of work which aim at learning probability distribution functions in various ways, discrete or continuous or both.

I hope that these new priorities may be extended beyond the CNCI program, so as to encourage more collaboration of engineers, computer scientists and other disciplines in finding unified approaches to these tasks.

## 3  Training of BLiSS Systems

There are two straightforward ways to try to move in the direction of BLiSS systems.

One way is to start from an existing design, like the particle swarm approach or Suykens' Fokker-Planck machine, and modify it to depend on **α**. For example, if the present design requires that you maintain a population of N choices of **u**, modify it to maintain N **networks** **u**[i](**α**), where i=1 to N. Then, on each iteration, instead of just changing **u**[i], **train u**[i](**α**), based on the same sorts of principles. Clearly there is a lot of room for experimentation and intuition here.

Another way is to use the new training methods (training **u**(**α**, **e** , T), where **e** is a vector of random numbers and T is a temperature parameter) which I propose in Werbos (1998b). This has the interesting implication that the parameter T needs to be adjusted over time, by the brain, as a function of circumstances. (For example, situations which require quick decision or high tension may call for a high adrenalin level and a low T, while relaxed situations which permit "brainstorming" may allow higher temperatures.) This fits well with models by Dan Levine and Sam Leven, which argued that variations in T (the level of "novelty seeking") is a crucial variable in explaining the fluctuations of human thought and behavior.

Of course, these two brief suggestions are not the whole story, and research is encouraged to fill out the story and explore many alternatives, in a mathematically grounded way.

## 4  Structure of OptionNets

The issue of structure will be very critical to achieving any kind of interesting performance here. For example – with the TSP or VLSI design problem, one would want to train networks to input problem descriptors of variable length. But ordinary neural networks involve fixed numbers of inputs and outputs! they do not have a rich enough structure to handle the full range of such problems – though they may be good enough for some preliminary reserach. Likewise, one would expect that an intelligent stochastic search would require the kind of iterative, relaxation approach that a Hopfield net (or other recurrent net) permits; ordinary feedforward networks would probably have very limited capabilities here.

Therefore, for maximum performance, research in this area will have to move relatively quickly to the use of more sophisticated structures or networks. Examples of such networks are the cellular SRN of Pang and Werbos (1998) and the Object Net design described in the "3 brain" paper cited in Werbos (1998) and in slides presented At ISAP99. (www.isap99.efei.br ) Again, however, these are simply the two possibilities which I happen to be most familiar with. NSF would be interested in all kinds of alternative structures at this time

In summary, the area of training and structuring Object Nets (or BLiSS systems in general) is almost a blank slate at present. There are many, many possibilities for researchers in this community to begin to fill in this slate, whihc in turn will be crucial to our hope of understanding and replicating the kind of intelligence we see in mammalian brains.

P. Werbos (1999), Neurocontrollers, in J. Webster, ed., *Encyclopedia of Electrical and Electronic Engineering*, Wiley.

P.Werbos (2009), Intelligence in the Brain: A theory of how it works and how to build it, *Neural Networks*, Volume 22, Issue 3, April 2009, Pages 200-212.

## Appendix. Larger Context and Update to 2010

The research challenge described in this paper is really just one step along the way to a much larger goal – the ability to understand and replicate the highest level of intelligence which exists even in the brain of the mouse. On the one hand, this task is a simplified reduced form of the design task which I called "Stochastic Action Nets" in Werbos (1998b). On the other hand, this is a difficult enough task in itself. In the previous decade, I awarded NSF grants to Wunsch (neural traveling salesman), to Serpen (simultaneous recurrent networks for stochastic search) and to Pelikan to begin to address this kind of challenge; however, they did not really succeed in addressing it head-on.

On May 24, 2010, Michael Fu of the University of Maryland and Barry Nelson led a new workshop for NSF on Simulation Optimization, http://users.iems.northwestern.edu/~nelsonb/workshop/, which revisited this topic. Russell Barton of Penn State University reviewed the work on "metamodels" or "forwards metamodels" -- the use of fast approximators to approximate the function $U(\mathbf{u},\mathbf{\alpha})$ as a function of $\mathbf{u}$, to assist in stochastic search across values of $\mathbf{u}$. He also mentioned his recent work in "inverse metamodeling" aimed at finding a metamodel to output a value of $\mathbf{u}$ which would meet some target outcome.

Even by 1998, Thaler had already had great success in stochastic search by training a classic kind of neural network (multilayer perceptron MLP gtrained by backpropagation) to serve as a "forwards global metamodel," a model of U across the entire space of possible choices $\mathbf{u}$. In the 2010 workshop, Barton noted that more researchers have been working with "local metamodels," metamodels of small regions in $\mathbf{u}$ space, because of the difficulty of obtaining global models, and because local metamodels are good enough for exploring the immediate neighborhood of the best value of $\mathbf{u}$ found so far; however, purely local metamodels are not useful in the larger task of finding the region of space where the best solution may be found, if U is not a convex function of $\mathbf{u}$.

In this workshop, I observed how the BliSS formulation here allows the formulation of a different kind of inverse metamodel, one which tries to map from $\mathbf{\alpha}$ to the optimal $\mathbf{u}$ for that $\mathbf{\alpha}$.

In the breakout group led by Jeffrey Herrmann of the University of Maryland, we groped for notation that would be more compelling to the operations research community to express this idea. In that community, it is common to say we are trying to maximize or minimize $f(\mathbf{x})$. To extend that notation, we may say that BliSS tries to maximize $f(\mathbf{x}_1,\mathbf{x}_2)$ as a function of $\mathbf{x}_2$. Or, as suggested by Hermann, we may use the letter "I" to denote an *instance* of the type of optimization problem we are trying to solve; in effect, "I" and $\mathbf{x}_2$" and "$\mathbf{\alpha}$" all represent the same idea.

It is exciting indeed that the research community may be ready to make a more serious effort to adress this kind of metamodeling. However, there is a key technical difficulty which needs to be addressed head-on. It is *easier* to learn local metamodels than to learn global metamodels, and it is easier to learn global metamodels than it is to learn inverse metamodels. There are two new technical challenges here: (1) how to choose a function approximator which is powerful enough to approximate a more challenging function (a function which is typically less smooth than the forwards function U and may involve many inputs over a large range of space); and (2) how to train the paramneters of that approximator, which may be challenging minimization problem in itself.

These technical difficulties are merely a special case of the larger challenge of learning predictive relations from data, which is a key part of the computatonal intelligence community. I have posted a current review of the opportunities and challenges here at //www.werbos.com/Neural/Erdos_talk_Werbos_final.pdf. (Unfortunately the file is too large to fit with current rules at arXiv.)

However, even a well trained inverse metamodel of this sort would still have some limitations. In many applications, one would not expect such an inverse metamodel to do as well as an expensive search dedicated to just one choice of $\mathbf{\alpha}$. However, such searches are usually very sensitive to their starting points. A trained inverse metamodel could be very useful in getting to the right area quickly, and in providing a "warm start" which reduces the time required to find the precise optimum. In some applications, the value of finding a good choice quickly may be greater than the value of finding the best choice over a long long time. With many neural network approximators, it is also possible to get full advantage of massively parallel chips such as Celular Neural Network chips, which allow even faster computation.

Once we start mastering such deterministic inverse metamodels, the next step up, closer to brain-like technology, is to build and use stochastic inverse metamodels, which learn to output new possibilities for $\mathbf{u}$ which have the best possible probability of finding a better choice than those now in hand, or which improve the quality of the metamodels now in hand.

That, in turn, is a step towards true Stochastic Action Networks. In Werbos (1998b), I stressed that there are really two types of "Action Network" in a complex, brain-like Approximate Adaptive Dynamic Programming (ADP) System. There are low-level action networks, which try to output vectors $\mathbf{u}(t)$ of current actions at time t. But there are also high-level action networks as part of "Decision Blocks," the kind of networks would address larger-scale choices such as where to go shopping today. High quality stochastic search is especially important and difficult at these higher levels. At the lowest level, in continuous time deterministic nonlinear ADP without jumps, S.N. Balakrishnan and J. Sarangapani have shown that the choice of **u**(t) is a trivial calculation, not requiring a network or metamodel at all, when the value function is available and we face a quadratic penalty on $\partial_t$**u**.

Werbos (2009) offers an overview, update and larger view than Werbos (1998b). It proposes that stochastic action networks for higher-level decisions are associated with a certain layer of the modern six-layer cerebral cortex, found in all mammals but not in lower vertebrates. Lower level brains like reptiles do have mechansims for developing complex maps of physical space. It is proposed that the brains capabilities with higher-level stochastic action networks are based on an adaptation of that cognitive mapping mechanism to the challenge of mapping the space of possible actions or decisions **u**. Perhaps we will have to learn to do likewise before we can build artificial systems with similar capabilities.